\documentclass[10pt,twocolumn,letterpaper]{article}

\usepackage[pagenumbers]{cvpr} 

\usepackage[dvipsnames]{xcolor}

\usepackage{indentfirst}
\usepackage{multirow}
\usepackage[accsupp]{axessibility} 

\definecolor{cvprblue}{rgb}{0.21,0.49,0.74}
\usepackage[pagebackref,breaklinks,colorlinks,citecolor=cvprblue]{hyperref}

\def\paperID{700} 
\def\confName{CVPR}
\def\confYear{2024}

\title{DiffAgent: Fast and Accurate Text-to-Image API Selection with Large Language Model}

\author{%
  Lirui Zhao$^{1,2\dagger\ddag}$ \quad Yue Yang$^{5,2\dagger}$ \quad Kaipeng Zhang$^{2\dagger\star}$ \quad Wenqi Shao$^{2\dagger\star}$ \\ Yuxin Zhang$^{1}$ \quad Yu Qiao$^2$ \quad Ping Luo$^{2,4}$ \quad Rongrong Ji$^{1,3*}$
  \\[0.2cm] 
  $^1$Key Laboratory of Multimedia Trusted Perception and Efficient Computing, \\ \; Ministry of Education of China, Xiamen University \\
  $^2$OpenGVLab, Shanghai AI Laboratory 
  $^4$The University of Hong Kong, \\
  $^3$Institute of Artificial Intelligence, Xiamen University
  $^5$Shanghai Jiao Tong University
}

\begin{document}

\maketitle

\renewcommand{\thefootnote}{\fnsymbol{footnote}}
{\let\thefootnote\relax\footnotetext{
\noindent \hspace{-5mm}$^*$Corresponding author: rrji@xmu.edu.cn\\
$^\dagger$Equal contribution\; $^\star$Project lead \\
$^\ddag$This work was done during his internship at Shanghai AI Laboratory.
}}

\begin{abstract}

Text-to-image (T2I) generative models have attracted significant attention and found extensive applications within and beyond academic research.
For example, the Civitai community, a platform for T2I innovation, currently hosts an impressive array of 74,492 distinct models.
However, this diversity presents a formidable challenge in selecting the most appropriate model and parameters, a process that typically requires numerous trials. 
Drawing inspiration from the tool usage research of large language models (LLMs), we introduce DiffAgent, an LLM agent designed to screen the accurate selection in seconds via API calls.
DiffAgent leverages a novel two-stage training framework, SFTA, enabling it to accurately align T2I API responses with user input in accordance with human preferences.
To train and evaluate DiffAgent's capabilities, we present DABench, a comprehensive dataset encompassing an extensive range of T2I APIs from the community.
Our evaluations reveal that DiffAgent not only excels in identifying the appropriate T2I API but also underscores the effectiveness of the SFTA training framework.
Codes are available at \url{https://github.com/OpenGVLab/DiffAgent}.
 
\end{abstract}
\section{Introduction}
\label{sec:introduction}

\begin{figure}[t]
  \centering
   \includegraphics[width=0.8\linewidth]{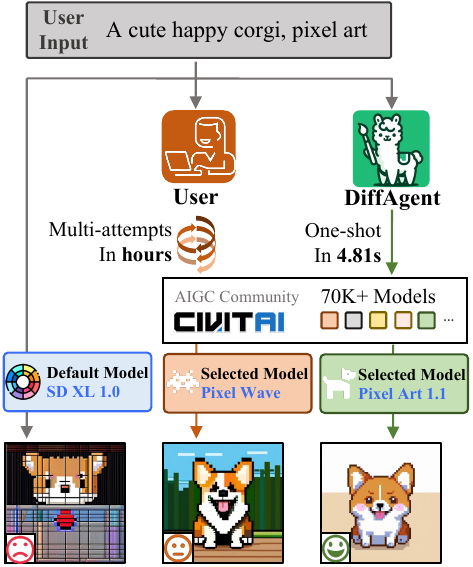}
   \vspace{-0.3cm}
   \caption{The comparison of different ways to T2I generation.}
   \label{fig:diffagent}
   \vspace{-0.7cm}
\end{figure}

Recent advancements in text-to-image (T2I) generative models~\cite{saharia2022photorealistic,nichol2022glide,ramesh2022hierarchical,rombach2022high,betker2023dalle3} have garnered considerable attention, courtesy of their extraordinary proficiency in following text prompts and the exceptional capability in image generation.
As a prominent technique, Stable Diffusion (SD)~\cite{rombach2022high} is widely utilized in various generative applications, and has gained significant traction within the research community and broader adoption.

However, the vanilla SD model falls short in catering to diverse personalized styles such as pixel art. For example, a corgi covered by several messy lines is generated by the SD-XL model~\cite{podell2023sdxl} with the user's request of `A cute happy corgi, pixel art', as shown in \cref{fig:diffagent}.
Such a limitation has spurred the emergence of various personalized models, which are tailor-made to excel in specific styles.
Innovations in lightweight personalization techniques such as Low-Rank Adaption (LoRA)~\cite{hu2021lora} and other approaches~\cite{gal2022image,ruiz2023dreambooth,yeh2023navigating} have streamlined the process of fine-tuning these models from the original SD framework, making it more cost-efficient.
Such advancements foster a wide adoption and influence of customized models in the open-sourced community~\cite{2016hugggingface,2022civitai}.
For instance, Civitai~\cite{2022civitai} hosts approximately 74,492 models with various personalized LoRA~\cite{hu2021lora} packages, facilitating itself a vibrant ecosystem for personalized image generation.
%


Despite the amounts of personalized models, it is hard to obtain the desired images given arbitrary requests from users. Firstly, it often requires multiple attempts to select the appropriate one from massive SD models. Secondly, continuous feedback is needed to modify generation parameters for the final performances.
The multiple attempts to select an appropriate model and refinement of input prompts necessitate a meticulous cycle of text-to-image (T2I) generations and continuous feedback adaptation, as shown in \cref{fig:diffagent}. This work targets tackling the challenging problem: \textit{how to quickly screen a good SD model and corresponding parameters for user-specific requests?}

Recent advances in large language models (LLMs) have shown that LLMs fine-tuned with self-instruction data about tool usage exhibit exceptional capability in invoking various tools, serving as powerful agents to call different tools (\eg, APIs) for target tasks. 
For example, Gorilla~\cite{patil2023gorilla} can generate proper model API collected from open-sourced model libraries such as HuggingFace~\cite{2016hugggingface} to execute various tasks. GPT4Tool~\cite{yang2023gpt4tools} enables LLMs to use multimodal tools for tackling a variety of visual problems such as object grounding and segmentation.
Motivated by these studies, we turn the challenges of selecting a proper SD model and corresponding parameters for various requests into a process of calling appropriate text-to-image APIs conditioned on the input prompt through LLMs.
In this regard, LLMs as API calling tools are expected to comprehend the user's request and generate a corresponding T2I API for image generation.
To achieve this, we propose LLM as the agent to call text-to-image API involving a stable diffusion model with associated hyperparameters in response to arbitrary user prompts, termed as DiffAgent.
DiffAgent incorporates two essential components: an instruction-following dataset called DABench that gathers extensive pairs of user prompts and T2I APIs, and a fine-tuning and alignment training framework for training an LLM as an agent to call a proper T2I API for the user prompt.
Compared with previous API calling tools \cite{patil2023gorilla,yang2023gpt4tools}, our DiffAgent has two appealing properties.
First, DiffAgent can screen the most appropriate T2I generation API for different user requests in the text-to-image task by training the LLM with massive pairs of user prompts and T2I APIs in DABench. While previous approaches~\cite{patil2023gorilla,yang2023gpt4tools} provide a limited number of model APIs for every single task. For example, GPT4Tools~\cite{yang2023gpt4tools} would always return Stable Diffusion XL as the API when given any text-to-image generation prompts.
Second, DiffAgent can generate the T2I API aligned well with human preferences. Other than supervised fine-tuning (SFT) LLM to follow API calling instructions as also done in previous approaches \cite{patil2023gorilla,yang2023gpt4tools}, DiffAgent further trains the LLM with the ranking responses with human feedback (RRHF) algorithm~\cite{yuan2023rrhf}, encouraging the LLM to generate responses aligned better with human preference than results from SFT.

Specifically, we first construct, DABench, a large instruction-following dataset for T2I generation API calling by crawling massive pairs of T2I APIs (SD models with corresponding parameters) and user prompts from the Civitai website which is a vibrant community for creators to share various AI-generated art~\cite{2022civitai}.
DABench undergoes rigorous filtering and reconstruction processes.
It then incorporates high-quality APIs from two main components of the community, \emph{i.e.} 39,670 and 10,812 pairs of T2I APIs and user prompt for SD 1.5 and SD XL, respectively, ultimately resulting in 50,482 pieces of instruction data for API calls. 

By fine-tuning the LLM (instantiated as LLaMA-2-7B \cite{touvron2023llama2} in this paper) on the DABench, DiffAgent can generate API calls that correspond to the user prompt.
Nevertheless, we find that the supervised fine-tuning on DABench cannot choose the best one in terms of human preference when multiple appropriate T2I APIs are presented for the user prompt.
As shown in \cref{fig:diffagent}, for one style (\eg, pixel art), there are several T2I generation APIs that all adopt this style but with different qualities.
To address this limitation, we further fine-tune the LLM with the RRHF algorithm~\cite{yuan2023rrhf} by incorporating various image assessment scores aligned with human preference as the reward model function.
Our observations indicate that the LLM, fine-tuned using the RRHF, exhibits a decreased occurrence of hallucination errors in its generated APIs (\eg, 20.1\% lower than the SFT model). Moreover, the resulting image aligns closely with human preferences. Specifically, when considering the SD 1.5 model architecture, our fine-tuned model demonstrates over a 10\% improvement compared to the SFT model in terms of unified metric as presented in Table \ref{tab:sd_results}.

This paper's contributions are summarized as follows:
\begin{itemize}[left=0.422cm]
    \item We propose DiffAgent, a text-to-image (T2I) agent model that selects appropriate T2I generation information (\ie, T2I API) to free users from tedious attempts and matching for the prompt.
    The T2I APIs from DiffAgent exhibit a high degree of human preferences and relevance with the prompt. 
    \item We introduced DABench, which is a carefully curated high-quality dataset for T2I API selection. DABench is the first API call dataset designed specifically for T2I domain, containing a large number of callable APIs (a total of 50,482).
    \item We conduct comprehensive experiments at DiffAgent, encompassing various datasets (DABench, COCO Caption~\cite{chen2015microsoft}, Parti Prompts\cite{yu2022scaling}) and multiple T2I evaluation scores (CLIP Score~\cite{hessel2021clipscore}, ImageReward~\cite{xu2023imagereward}, HPS v2~\cite{wu2023human}).
    Our results demonstrate that DiffAgent achieves a substantial performance improvement than using the original T2I model (\eg, 40\% in SD 1.5).
\end{itemize}

\section{Related Work}
\label{sec:related_work}

\textbf{LLMs Tool Usage. }
Recent studies~\cite{patil2023gorilla,qin2023toolllm,yang2023gpt4tools,tang2023toolalpaca,schick2023toolformer,hao2023toolkengpt} have revealed the growing potential of large language models (LLMs) in acquiring proficiency in tool usage and decision-making within intricate environments.
\cite{schick2023toolformer,hao2023toolkengpt} enhance the performance of LLMs in targeted domains by using a small set of simple tools (\eg, using calculators for precise mathematical reasoning).
Gorilla~\cite{patil2023gorilla} enables LLMs to respond to API calls by constructing a dataset and performing fine-tuning. 
ToolLLM~\cite{qin2023toolllm} further to make LLMs utilize massive APIs.
GPT4Tool~\cite{yang2023gpt4tools} focuses on using multimodal tools for tackling a variety of visual problems (\eg, segmentation).
ToolAlpaca~\cite{tang2023toolalpaca} verifies the feasibility of equipping compact LLMs with generalized tool-use capacities
These works have primarily focused on a general range of API calls that only allow for limited and fixed options for a single task.
Our work, DiffAgent, aims to select the most suitable API from massive T2I APIs in the text-to-image task.
Different further align LLMs with human preferences using the training framework SFTA.

\begin{figure*}[t]
  \centering
   \includegraphics[width=1\linewidth]{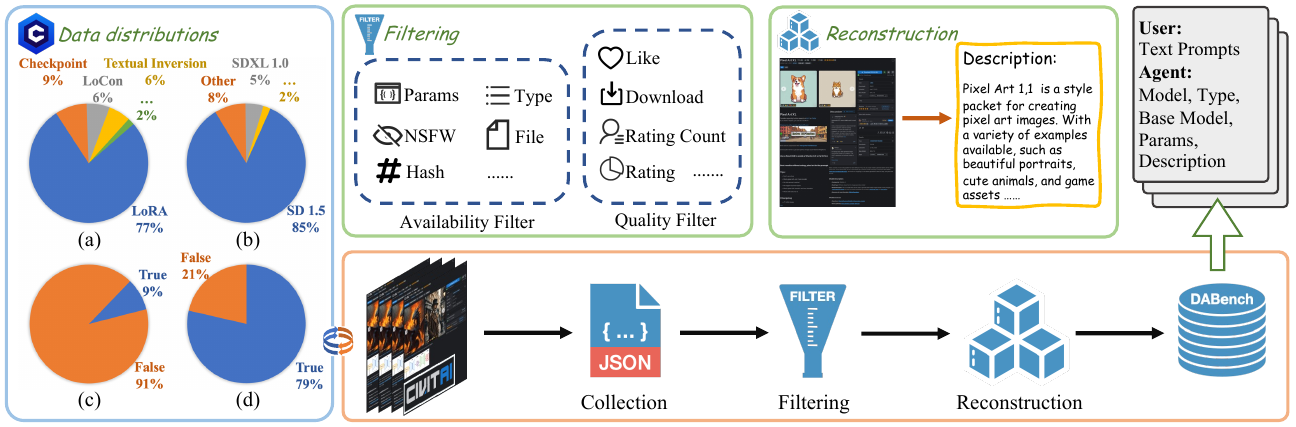}
   \caption{The data collection process of DABench. The left side is the data distributions of the data source, Civitai~\cite{2022civitai}. These distributions include: (a) Model type. (b) Base model architecture. (c) Not suitable for work (NSFW) content. (d) File availability.}
   \label{fig:data_collection}
   \vspace{-0.5cm}
\end{figure*}
\textbf{Text-to-Image Diffusion Models. }
Recently, text-to-image (T2I) diffusion models~\cite{nichol2022glide,ramesh2022hierarchical,betker2023dalle3}, have shown exceptional capability in image generation quality and extraordinary proficiency in following text prompts, under the dual support of large-scale text-image dataset~\cite{schuhmann2022laion} and model optimizations~\cite{dhariwal2021diffusion,ho2020denoising,rombach2022high}.
GLIDE~\cite{nichol2022glide} incorporated text conditions into the diffusion model and empirically showed that leveraging classifier guidance leads to visually appealing outcomes.
DALLE-2~\cite{ramesh2022hierarchical} enhances text-image alignment via CLIP~\cite{radford2021learning} joint feature space. 
DALLE-3~\cite{betker2023dalle3} further improves the prompt following abilities by training on highly descriptive generated image captions.
Stable Diffusion~\cite{rombach2022high}, a well-established and widely adopted method, has gained substantial traction in the open-source community~\cite{2022civitai,2016hugggingface}, leading to a large and diverse model library (\eg, 74492 models in Civitai~\cite{2022civitai}).
The cost of generating desired images using specific personalized models also increases especially in such a vast quantity of models with quality varies.
DiffAgent can free users from multiple attempts and screen the personalized model and corresponding parameters for matching text prompts in seconds.

\textbf{Align Language Models with Human Preferences. }
Reinforcement Learning from Human Feedback (RLHF) has been pivotal in aligning LLMs to human preferences, typically employing Proximal Policy Optimization (PPO) algorithm for model optimization~\cite{stiennon2020learning,ziegler2019fine,ouyang2022training}. 
During the optimization process with PPO, the likelihood of a whole generation is required to update LLMs.
For LLM agents, however, human feedback is typically available only after acquiring the complete API response and successful function invocation.
This implies that 1) the reward model based on human feedback is not capable of getting the likelihood of a whole generation and 2) the receipt of delayed human feedback necessitates the complete API response for effective invocation.
Rank Responses to Align Human Feedback (RRHF)~\cite{yuan2023rrhf} emerges as a notable method, employing reward models to rank multiple responses for language model alignment.
This approach facilitates its extension to fine-grained API Agents, maximizing the utility of existing reward models.
Nevertheless, it presents challenges such as maintaining the output API’s integrity and evaluating output using the reward model solely on the post-invocation actual results.
To address these challenges, we enhance the RRHF method by incorporating evaluation scores into the reward model function, establishing a viable alignment procedure.

\section{Methodology}
\label{sec:methodology}

Given a text prompt, our objective is to free users from tedious attempts and matching in order to find the appropriate APIs for text-to-image generation (\ie, T2I API selection).
We initially introduce DABench for this task, a benchmark derived from Civitai.
DABench consists of Instruction-API pairs, each pair including the text prompts and corresponding T2I generation information (\eg, model information and parameters information).
We provide an overview of the procedure of the API dataset collection, filtering, and reconstruction as detailed in \cref{sec:dabench}.
Subsequently, we introduce DiffAgent, an LLM agent to call T2I APIs in response to arbitrary user prompts.
We describe the multi-step training framework, called SFTA, for DiffAgent and the unified metric for evaluating DiffAgent (\cref{sec:diffagent}).
%

\subsection{DABench}
\label{sec:dabench}
The Civitai community~\cite{2022civitai} comprises approximately 74,492 models, each typically accompanied by a set of sample images.
These images demonstrate the function of the corresponding model, while their prompts show the style and common scenarios that are most suitable for generating.
Civitai also exists evaluation mechanisms (download count, rating, \etc), statistical results show common preferences of the community.
We can acquire hierarchical information, which includes the model's data and sample images with generation details from uploader, and statistical results from user.
Our goal is to collect a high-quality dataset, each data pair consists of Instruction (text prompts), and its API (corresponding model and parameter information).
This is achieved through the collection and filtration of model-level data, followed by a reverse mapping with image's data.

\paragraph{APIs Collection}
\label{sec:api_collection}
We first perform a comprehensive analysis of the composition of Civitai.
As shown in \cref{fig:data_collection}, the base model architectures can be categorized into two primary types: SD 1.5 ($85\%$) and SDXL 1.0 ($5\%$).
regarding the model types, they consist of LoRA ($77\%$) and Checkpoint ($9\%$).
Therefore, we determine the scope of APIs collection to include the above content.

\paragraph{APIs Filtering}
Due to the varying quality of APIs, we employ two multi-attribute filters (availability and quality filters) to choose a wide range of high-quality options.

As depicted in \cref{fig:data_collection}, we initiate the process of verifying the availability of APIs to ensure successful invocation of the collected APIs.
The APIs related to not suitable for work (NSFW) content will also be excluded during this stage.
As stated in \ref{sec:api_collection}, the model type (Checkpoint and LoRA) and base model architectures (SDXL 1.0 and SD 1.5) under the scope are reserved and vice versa.
During the T2I generation, the LoRA model needs to be used simultaneously with the base model by embedding tags into prompts.
However, only the hash of the base model exists in the collected data, and for a LoRA model, there may be different base models present in different sample images.
Then we attempt to obtain base model information for LoRA models' images using the hash and filter out the inaccessible parts.
The LoRA tag embedded within their text prompts is subsequently eliminated.

We next perform a quality assessment of these APIs using established evaluation mechanisms (download count, rating count, \etc) within the community, showing the common preferences of the users.
The process results in the identification of a subset with superior quality

\paragraph{Description Reconstrcution}
As an active open-source community, many models lack accurate and relevant descriptions.
For instance, the description of the pixel Art model does not include many of the applicable styles for the model itself while we can understand this at a glance through sample images.
Hence we employ GPT-3.5-turbo~\cite{2022chatgpt} to reconstruct the description of models by incorporating information from both the models themselves and sample images (utilizing their associated prompts) as shown in \cref{fig:data_collection}.
This approach aims to improve the consistency between APIs and their corresponding description.

\paragraph{Dataset composition}
We convert the collected APIs' information into a json object with the following fields: \{model, type, base Model, width, height, sampling\_method, sampling\_steps, cfg\_scale, model\_description\}.
These fields consist of two parts required for T2I generation: model information (model, type, base Model, model\_description) and parameter information (the remaining fields).
The collected instructions are also converted into a json object: \{prompt, negative\_prompt\}.

Finally, DABench includes high-quality APIs from SD 1.5 (39,670 Instruction-API pairs in 5,516 API calls) and SD XL (10,812 in 1,306).
%

\begin{figure*}[t]
  \centering
   \includegraphics[width=0.9\linewidth]{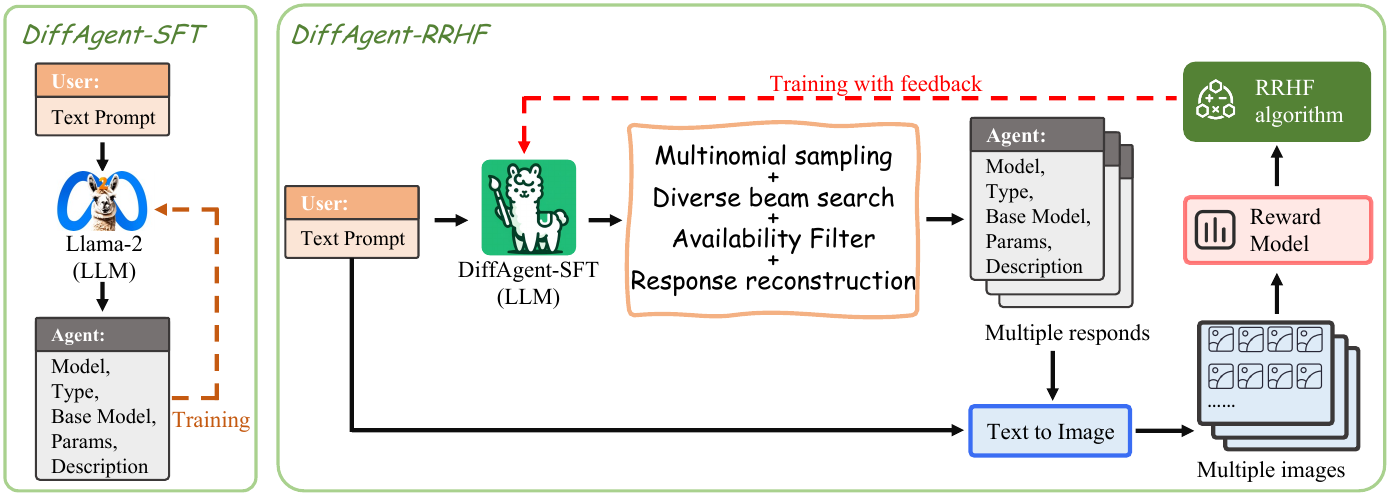}
   \caption{The training framework SFTA to get DiffAgent.}
   \label{fig:framework}
   \vspace{-0.5cm}
\end{figure*}

\subsection{DiffAgent}
\label{sec:diffagent}
DiffAgent is an LLM agent fine-tuned through our two-stage training framework, SFTA, to achieve T2I API selection given by text prompt.
In the first stage, we employ supervised fine-tuning (SFT) with DABench to train and obtain the preliminary model known as DiffAgent-SFT.
In the second stage, we further enhance DiffAgent-SFT through alignment with human preferences as depicted in \cref{fig:framework}, to obtain the final DiffAgent model (\ie, DiffAgent-RRHF).

\subsubsection{DiffAgent-SFT}
We employ the standard fine-tuning to get the DiffAgent-SFT model.
For a pair of text prompts $t$ and API response $r$, the cross-entropy loss is used for fine-tuning:
\begin{equation}
    L_{sft} = -\sum_t \log P_\pi(r_{t}|t,r_{<t}).
\end{equation}
DiffAgent-SFT possesses all the necessary conditions for generating images, including text prompts as input along with the corresponding model and parameters as response API.
This enables us to generate images directly for the purpose of evaluating the performance of APIs.

\subsubsection{DiffAgent-RRHF}
We first propose a human preference-related unified metric for evaluating DiffAgent. Then, we introduce an effective training framework to further align DiffAgent-SFT with human preferences.

\paragraph{Human Preferences Evaluation}
\label{human_preferences_evaluation}

In the T2I generation domain, there already exist multiple scoring models or metrics~\cite{wu2023human,xu2023imagereward,hessel2021clipscore}, which can be employed for related or human preference evaluation.
Given a prompt $t$ in text prompts and the generated image $x\left(r\right)$ using API response $r$, we evaluate DiffAgent using a human preference-related unified metric composed of:
\textit{First}, CLIP Score~\cite{hessel2021clipscore} aims to evaluate the correlation between $t$ and $x$ using cosine similarity of their respective embedding obtained through CLIP~\cite{radford2021learning}:
\begin{equation}
  s_{\theta_c}(t, x\left(r\right)) = w * \max\left(\cos (\mathrm{Enc}_{\rm txt}(t), \mathrm{Enc}_{\rm img}(x\left(r\right))), 0\right),
  \label{eq:clipscore}
\end{equation}
where $w=2.5$ as stated in \cite{hessel2021clipscore}, and $\theta_c$ represents the parameters in CLIP (ViT-B for image encoder, 12-layers transformer for text encoder).
\textit{Second}, ImageReward~\cite{xu2023imagereward}, a T2I human preference reward model fine-tuned by BLIP~\cite{li2022blip} (ViT-L for image encoder, 12-layers transformer for image-grounded text encoder), extracts $t$ and $x\left(r\right)$ features, combine them with cross attention, and employs an MLP to generate a scalar:
\begin{equation}
  s_{\theta_i}(t, x\left(r\right)) = \mathrm{MLP}(\mathrm{Enc}_{\mathrm{txt}}(t, \mathrm{Enc}_{\mathrm{img}}(x\left(r\right)))),
  \label{eq:imagereward}
\end{equation}
where $\theta_i$ represents the parameters of the ImageReward Model.
\textit{Third}, Human Preference Score v2 (HPS v2)~\cite{wu2023human}, a scoring model fine-tuned by CLIP-H model, can more accurately predict text-generated images' human preferences:
\begin{equation}
  s_{\theta_h}(t, x\left(r\right)) = \frac{\mathrm{Enc}_{\rm txt}(t) \cdot \mathrm{Enc}_{\rm img}(x\left(r\right))}{\tau},
  \label{eq:hps}
\end{equation}
where $\tau$ represents the learned temperature scalar during the fine-tuning process of HPS v2, and $\theta_h$ denotes the parameters of HPS v2.

Now, we will proceed to the normalization and combination of the scores to obtain a unified metric $S$ for evaluating APIs.
Given a text prompt $t$ and a set of images $X\left(r\right)=\left\{x\left(r\right)_1,x\left(r\right)_2,...,x\left(r\right)_k\right\}$, where each image $x\left(r\right)_i$ generated based on a different random seed to mitigate the influence of random seeds, we compute the scores from three metrics using \cref{eq:clipscore,eq:imagereward,eq:hps}.
Each of these scores is normalized to the range $[0,1]$.
%
The final score for the text prompt $t$ with API response from DiffAgent is obtained by averaging all image's scores:
\begin{equation}
  S(t,r) = \frac{1}{k} \sum_{i=1}^{k} S(t, x\left(r\right)_i),
  \label{eq:score}
\end{equation}
\text{s.t.}
\begin{equation}
\begin{aligned}
    S(t, x\left(r\right)_i) = &\frac{1}{3}
      \left.( s_{\theta_c, \text{norm}}(t, x\left(r\right)_i) + \right.\\
      &\left. s_{\theta_i, \text{norm}}(t, x\left(r\right)_i) + s_{\theta_h, \text{norm}}(t, x\left(r\right)_i) \right.).
\end{aligned}
\end{equation}
The Unified Metric $S(t,r)$ describes the human preference score for an API response $r$ given a prompt $t$.
The \cref{eq:score} serves a dual purpose, as it can be utilized for both API evaluation and as a reward model function in alignment with human preferences.

\paragraph{Alignment with Human Preferences}

The existing Reinforcement Learning from Human Feedback (RLHF) methods~\cite{ouyang2022training,christiano2017deep,stiennon2020learning} is performed using reinforcement learning algorithms (\eg, PPO~\cite{schulman2017proximal}), which depends on the likelihood of a whole generation to update the model.
However, the evaluation way of the agent model necessitates the exclusion of vanilla RLHF methods due to the delayed feedback received only upon API invocation and a complete API response is required during this process.

We propose an extension to the existing RRHF~\cite{yuan2023rrhf} method that can be used for aligning DiffAgent (\cref{fig:framework}).
For text prompts $t$, we employ multinomial sampling to get a comprehensive API response and utilize diverse beam search~\cite{vijayakumar2016diverse} to get $m$ diverse responses from SFT model $\pi$.
However, it is difficult to ensure the format and validity of multiple different responses.
Therefore, we only take the model information part of responses, then  proceed to validate and reconstruct the complete API response information.
After incorporating the default API response, a list of $n$ API responses $R=\left\{r_1,r_2,...,r_n\right\}$ is generated.
Then we actually call APIs to retrieve the images and use \cref{eq:score} as the scoring function for the reward model to get $n$ scores for each $r_i$ with $S(t,r_i)=s_i$.

To align with scores $\left\{s_i\right\}_n$, we use SFT model $\pi$ to give scores $p_i$ for each $r_i$ by:
\begin{equation}
    p_i = \frac{\sum_t \log P_\pi(r_{i,t}|t,r_{i,<t})}{\|r_i\|},
\end{equation}
where $p_i$ is conditional log probability (length-normalized) of $r_i$ under model $\pi$.
We use the RRHF algorithm to let the model $\pi$ give larger probabilities for better responses and give smaller probabilities for worse responses.
this object is optimized by ranking loss:
\begin{equation}
    L_{rank} = \sum_{s_i<s_j}\max(0,p_i-p_j),
\end{equation}
and a cross-entropy loss like SFT process is added as well:
\begin{equation}
    L_{ce} = -\sum_t \log P_\pi(r_{i',t}|t,r_{i',<t})
\end{equation}
\text{s.t.}
\begin{equation}
    i' = \arg\max_i{s_i}.
\end{equation}
The overall loss is utilized to optimize the DiffAgent-SFT in order to obtain DiffAgent-RRHF:
\begin{equation}
    L=L_{rank}+L_{ce}.
\end{equation}

\section{Experiment}
\label{sec:experiment}

We performed a comprehensive evaluation of our model, DiffAgent, by conducting experiments on three benchmark datasets: DABench, COCO Caption~\cite{chen2015microsoft}, and Parti Prompts~\cite{yu2022scaling}.
Our evaluation involved comparing the performance of DiffAgent with other powerful models and investigating the effectiveness of our two-stage training framework.
In addition, we show the effectiveness of aligning with human preferences in DiffAgent through comparative experiments.
We then conducted a user study to examine the ability of DiffAgent to identify related diverse styles' APIs, which demonstrated its strong capability in this regard.
Finally, we showcase visualizations of our generated images to offer an intuitive assessment of Diffagent's capabilities and facilitate comparisons.

\begin{table}[!t]
  \centering
  \small
    \begin{tabular}{llcc}
    \toprule
        DiffAgent   &                           & SFT    & RRHF              \\ \midrule
        \multirow{3}{*}{SD 1.5} 
        & DABench                               & 11.5   & \textbf{9.5}      \\
        & COCO Caption~\cite{chen2015microsoft} & 26.9   & \textbf{15.5}     \\
        & Parti Prompts~\cite{yu2022scaling}    & 28.9   & \textbf{20.0}     \\ \midrule
        \multirow{3}{*}{SD XL}  
        & DABench                               & 7.4    & \textbf{7.1}      \\
        & COCO Caption~\cite{chen2015microsoft} & 45.6   & \textbf{36.0}     \\
        & Parti Prompts~\cite{yu2022scaling}    & 48.1   & \textbf{28.0}     \\ \bottomrule
    \end{tabular}
  \caption{The hallucination error ($\downarrow$) of DiffAgent in different training stage. The DiffAgent-RRHF shows a lower error than DiffAgent-SFT.}
  \label{tab:hallucination}
  \vspace{-0.5cm}
\end{table}

\begin{table*}[!ht]
  \centering
  \captionsetup{skip=5pt}
  \small
    \begin{tabular}{llcccc|cccc}
    \toprule
      &  & \multicolumn{4}{c|}{\textbf{SD XL}} & \multicolumn{4}{c}{\textbf{SD 1.5}} \\
     && \begin{tabular}[c]{@{}c@{}}CLIP\\Score $\uparrow$\end{tabular}      & \begin{tabular}[c]{@{}c@{}}Image\\Reward $\uparrow$\end{tabular}  & \begin{tabular}[c]{@{}c@{}} HPS\\v2 $\uparrow$ \end{tabular}        & \begin{tabular}[c]{@{}c@{}} Unified\\Metric $\uparrow$ \end{tabular}& \begin{tabular}[c]{@{}c@{}}CLIP\\Score $\uparrow$\end{tabular}      & \begin{tabular}[c]{@{}c@{}}Image\\Reward $\uparrow$\end{tabular}  & \begin{tabular}[c]{@{}c@{}} HPS\\v2 $\uparrow$ \end{tabular}        & \begin{tabular}[c]{@{}c@{}} Unified\\Metric $\uparrow$ \end{tabular}  \\
    \midrule
        \multirow{5}{*}{DABench}   
        &Baseline           & 83.2          & 65.1          & 26.8          & 49.2     &75.4           &-73.9          &25.6           &37.2       \\
        &Baseline*          & \textbf{83.4} & 67.4          & 26.9          & 54.2      &75.4           &-69.7          &25.7           &39.5       \\
        &DiffAgent\#      & 83.0          & 69.6          & 26.9          & 53.0   &80.6           &-6.7           &26.6           &67.0         \\
        &DiffAgent-SFT   & 83.2          & 67.7          & 27.0          & 55.8     &79.2           &-21.0          &26.4           &59.7       \\
        &DiffAgent-RRHF  & 83.2          & \textbf{71.4} & \textbf{27.0} & \textbf{57.5}   &\textbf{80.6}  &\textbf{-5.3}  &\textbf{26.6}  &\textbf{68.2} \\
    \midrule
        \multirow{5}{*}{COCO Caption~\cite{chen2015microsoft}} 
        &Baseline                     & 85.2          & 71.8          & 27.4          & 48.0 &78.8           &-37.6          &26.3           &42.3\\
        &Baseline*                    & \textbf{85.5} & 75.3          & 27.6          & 53.7 &78.8           &-33.0          &26.4           &45.2\\
        &DiffAgent\#                & 85.0          & 78.9          & 27.5          & 52.2&80.3           &7.4            &26.9           &60.8 \\
        &DiffAgent-SFT             & 85.4          & 79.2          & 27.6          & 55.6 &79.5           &-12.4          &26.6           &53.5\\
        &DiffAgent-RRHF            & 85.3          & \textbf{82.7} & \textbf{27.6} & \textbf{57.4} &\textbf{80.3}  &\textbf{9.6}   &\textbf{26.9}  &\textbf{62.2}\\
    \midrule
        \multirow{5}{*}{Parti Prompts~\cite{yu2022scaling}}

        &Baseline                     & 86.5          & 88.7          & 27.4          & 49.8 &79.9           &-30.1          &26.1           &43.0 \\
        &Baseline*                    & 86.6          & 91.3          & 27.4          & 53.6&80.2           &-24.6          &26.2           &46.5 \\
        &DiffAgent\#                & 86.4          & 93.5          & 27.4          & 52.5 &81.6           &12.3           &26.7           &60.7\\
        &DiffAgent-SFT             & 86.7          & 96.4          & 27.6          & 56.5&80.3           &-0.2            &26.5           &53.7  \\
        &DiffAgent-RRHF            & \textbf{86.8} & \textbf{98.6} & \textbf{27.6} & \textbf{58.8} &\textbf{81.8}  &\textbf{13.2}  &\textbf{26.7}  &\textbf{61.8}\\
    \bottomrule
    \end{tabular}
  \caption{The evaluation results in SD XL and SD 1.5 model architecture. Baseline*: Use the baseline model with the parameter information from DiffAgent-RRHF for generation. DiffAgent\#: Use the model information from DiffAgent-RRHF with default parameter}
  \label{tab:sd_results}
  \vspace{-0.5cm}
\end{table*}

\subsection{Experimental Settings}

\paragraph{Training Setups}
We transform DABench's Instruction-API pairs into single-round user-agent dialogues, partitioning them in an 8:1:1 ratio for fine-tuning, aligning with human preferences, and evaluation.
For different model architectures (SD 1.5 and SD XL), we employ the SFTA framework to fine-tune LLaMA-2-7b~\cite{touvron2023llama2}, thereby generating the respective DiffAgents.
In the SFTA's second stage, we assess each response through ten random image generations ($k=10$, as per \cref{eq:score}).

\paragraph{Dataset Setings}
DiffAgent's evaluation employs the following datasets:
1) The validation set of DABench.
2) For the COCO Caption and Parti Prompts datasets, we randomly select 1,000 prompts from each for assessment. Additionally, we adapt GPT-3.5-turbo~\cite{2022chatgpt} as a text prompt generator, leveraging in-context learning to transform concise prompts into well-structured text prompts.

\paragraph{Baselines}
Our primary comparison involves DiffAgent against the original Stable Diffusion (SD) with a default parameter, maintaining identical architectures.
Given that DiffAgent's T2I APIs incorporate both model and parameter information, we consider combining these two parts separately with the baseline to obtain two additional comparison objects.
Furthermore, the DiffAgent-SFT, derived from the initial stage of the SFTA process, serves as a significant method in our analysis.
As outlined in \cref{human_preferences_evaluation}, we employ three metrics (CLIP Score~\cite{hessel2021clipscore}, ImageReward~\cite{xu2023imagereward}, and HPS v2~\cite{wu2023human}) alongside the Unified Metric (\cref{eq:score}) for evaluation.
The Unified Metric requires the normalization of scores across different baselines, a process complicated by potential hallucination issues in DiffAgent's generated API.
Consequently, we consider applying the baseline to generate images when the issue of hallucination arises.
we standardized the shape of the generated image to be consistent due to the impact of shape on the evaluation.
The default parameters are set to \{sampling\_method: Euler a, sampling\_steps: 20, cfg\_scalse: 7\}, which are the most commonly used parameter choices.

\subsection{Results}

\subsubsection{Hallucination}
In Table \ref{tab:hallucination}, we present the hallucination error of DiffAgent on various evaluation datasets.
It is evident that DiffAgent-RRHF consistently achieves a lower hallucination error compared to DiffAgent-SFT.
For example, The hallucination error decreased from 48.1\% to 28.0\% after optimizing with RRHF on the Parti Prompt.
This proves that the RRHF stage of our framework SFTA can reduce the hallucination error and provide more callable API responses.

\subsubsection{Main Experiment}
\cref{tab:sd_results} shows the main results of DiffAgent with different baselines. We report the results of three T2I metrics and the Unified Metric by averaging the scores of three randomly generated images for each T2I API.
We can deduce the following conclusions from the results in several aspects.
\paragraph{Unified Score of DiffAgent}
The DiffAgent significantly outperforms the baseline in the Unified Metric in all datasets.
Remarkably, DiffAgent-RRHF surpasses the baseline by 8.3 $ \sim $ 9.4 in the SD XL architecture and 18.8 $ \sim $ 31 in SD 1.5.
We find that this improvement is mainly due to the contribution of human preference scores (ImageReward and HPS v2), which indicates that DiffAgent can generate T2I APIs that align well with human preferences.
Notably, the models from T2I API are typically specialized for specific styles, resulting in similar outcomes for Clip Score compared with baseline.
At the same time, we can observe that under the same architecture, the Unified Scores are very close across different datasets. This proves that this score is a powerful and robust evaluation metric for T2I generation.

\begin{figure*}[!t]
  \centering
   \includegraphics[width=0.9\linewidth]{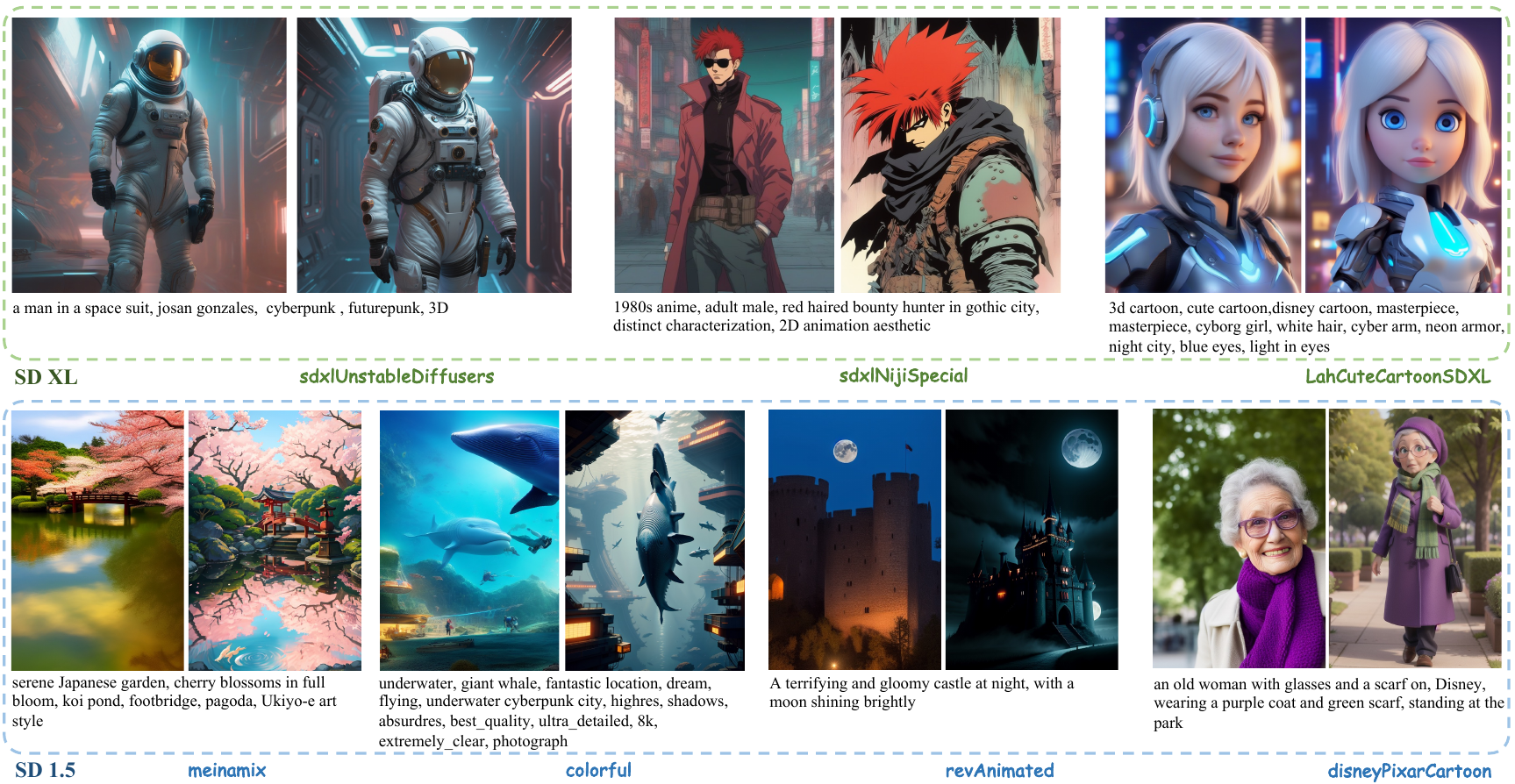}
   \caption{The visualization comparison between the original SD model (left) with DiffAgent's T2I API (right). The two lines come from the SD XL and SD 1.5 architectures respectively. For each pair, we provide the user prompt and the model name from T2I API.}
   \label{fig:visualization}
   \vspace{-0.5cm}
\end{figure*}
\paragraph{Effectiveness of Different Components}
The Baseline* and DiffAgent\# respectively demonstrate the utilization of either parameter or model information from DiffAgent independently, resulting in improved performance compared to the baseline.
This shows these two components of T2I API both have a positive effect on original T2I generation.
Furthermore, we observe that the model information from API exhibits a greater improvement for SD 1.5 architecture, whereas, for XL architecture, there is from parameter information.
This observation can be attributed to the thriving popularity of models under the SD 1.5 architecture within the community, as depicted in Figure \ref{fig:data_collection}.

\paragraph{Effectiveness of RRHF}
We also report the evaluation of DiffAgent-SFT, which represents the results in the first stage of our training framework SFTA, to compare with DiffAgent-RRHF
As shown in \cref{tab:sd_results}, the DiffAgent-RRHF's performance is higher than the DiffAgent-SFT, especially in human preference scores (\eg, in SD 1.5 architecture, 15.7 improvement in DABench using ImageReward). 
Notably, DiffAgent-RRHF markedly enhances the unified metric scores by 8.1 $ \sim $ 8.7 than DiffAgent-SFT under SD 1.5.
This indicates that the RRHF process improves the alignment of responded API with human preferences.

\paragraph{Inference Time}
We evaluate the average time overhead of API generation on DABench, which is 4.81s (single A100 80G), by processing each data independently.
This showcases the superior efficient applicability of DiffAgent.
%

\begin{figure}[!b]
  \centering
   \vspace{-0.5cm}
   \includegraphics[width=0.8\linewidth]{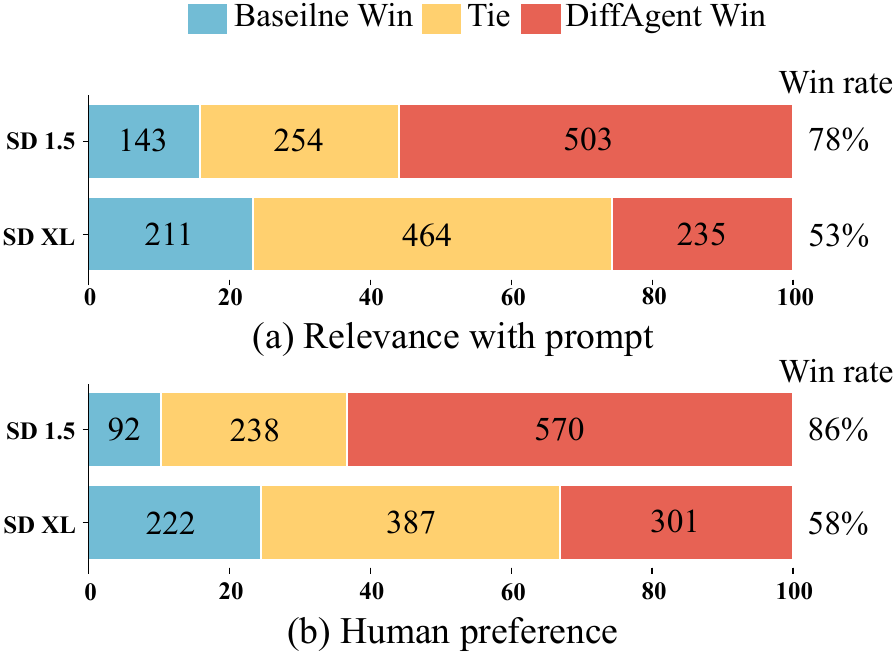}
   
   \caption{The user study results of DiffAgent. Win rates are calculated without considering tie samples. DiffAgent surpasses the baseline in both relevance with prompt and human preference.}
   \label{fig:user_study}
\end{figure}

\subsection{Human evaluations}
We further evaluate our DiffAgent using a user study.
Users are provided with two images, one from DiffAgent and one from baseline, along with a corresponding prompt.
Users can choose which is better or a tie for the relevance with prompt and image quality (as human preferences).
Here, under different model architectures, we sample respective 90 image pairs and prompts evenly across the three evaluation datasets (DABench, COCO Caption, and Parti Prompts) and received evaluations from ten users.
As shown in \cref{fig:user_study}, the user study results align with the unified metric and demonstrate that DiffAgent shows more relevance and human preferences than Baseline.
Specifically, DiffAgent achieves a win rate of  78\% on relevance and 86\% on human preference compared to the baseline in SD 1.5, while for SD XL, it achieves a win rate of 53\% and 58\%.
%

\subsection{Visualizations}
\label{sec:visualizations}
We present a visual comparison between the baseline and T2I API from DiffAgent in \cref{fig:visualization}.
The first row is the results under the SD XL architecture, and the second row is under the SD 1.5.
From the comparison, it can be seen that compared to the original one, DiffAgent can find T2I APIs that suit the style of the prompts.
For example, the DiffAgent can capture the Disney style in prompts and responses a T2I API includes the "disneyPixarCartoon" model, a model to make pictures maintain a Disney style.

\section{Conclusion}
\label{sec:conclusion}

Our research is centered on harnessing LLMs as agents to liberate users from constant attempts and adjustments of the generative information in the text-to-image (T2I) domain.
This paper introduces DiffAgent, a novel agent model adept at generating T2I APIs (model and parameter information) in response to user specifications.
Additionally, we present a specialized training framework designed to fine-tune LLMs, enabling them to adeptly select APIs that align well with human preferences.
Massive evaluations demonstrate that DiffAgent markedly outperforms existing powerful models, such as SD XL, in both our collected dataset and two widely used caption datasets. Notably, DiffAgent exhibits a remarkable proficiency in swiftly identifying the most suitable T2I APIs, expanding the tool usage of LLM in specific domains.

\section*{Acknowledgement}
This work was supported by National Science and Technology Major Project (No. 2022ZD0118202), the National Science Fund for Distinguished Young Scholars (No.62025603), the National Natural Science Foundation of China (No. U21B2037, No. U22B2051, No. 62176222, No. 62176223, No. 62176226, No. 62072386, No. 62072387, No. 62072389, No. 62002305 and No. 62272401), and the Natural Science Foundation of Fujian Province of China (No.2021J01002,  No.2022J06001).

{
    \small
    \bibliographystyle{ieeenat_fullname}
    \bibliography{main}
}


\end{document}